\documentclass[sigconf]{acmart}

\AtBeginDocument{%
  \providecommand\BibTeX{{%
    \normalfont B\kern-0.5em{\scshape i\kern-0.25em b}\kern-0.8em\TeX}}}

\renewcommand\footnotetextcopyrightpermission[1]{} 

\usepackage{adjustbox}
\usepackage[flushleft]{threeparttable}
\usepackage{hyperref}
\usepackage{stackengine}
\usepackage{colortbl}
\usepackage{caption}
\usepackage{microtype}
\usepackage{tabularx}
\usepackage{xcolor}
\usepackage{subfigure}
\usepackage{multirow}
\usepackage{booktabs}
\usepackage{multirow}

\usepackage{geometry}
\usepackage{algpseudocode}
\usepackage{algorithm}
\usepackage{subfigure}
\usepackage{caption}
\usepackage[many]{tcolorbox}
\usepackage{multirow}
\usepackage{booktabs}
\usepackage[labelformat=parens,labelsep=quad,skip=3pt]{caption}
\usepackage{graphicx}
\usepackage[T1]{fontenc}
\usepackage[super]{nth}
\usepackage{amsmath}

\begin{document}

\title{
DeepKnowledge: Generalisation-Driven Deep Learning Testing }




\author{Sondess Missaoui}
\email{sondess.missaoui@york.ac.uk}
\affiliation{%
  \institution{Department of Computer Science, University of York, UK}
}

\author{Simos Gerasimou}
\email{simos.gerasimou@york.ac.uk}
\affiliation{%
  \institution{Department of Computer Science, University of York, UK}
}

\author{Nikolas Matragkas}
\email{nikolaos.matragkas@cea.fr}
\affiliation{%
  \institution{CEA List,  Universit\'e Paris-Saclay, France}
}

\newcommand{\approach}{\textsc{DeepKnowledge}}
\newcommand{\sm}[1]{\textcolor{cyan}{[#1]}}
\newcommand{\simos}[1]{\textcolor{magenta}{[#1]}}
\newcommand{\remove}[1]{\textcolor{gray!30}{[#1]}}

\begin{abstract}
Despite their unprecedented success, DNNs are notoriously fragile to small shifts in data distribution, demanding effective testing techniques that can assess their dependability. Despite recent advances in DNN testing, there is a lack of systematic testing approaches that assess the DNN’s capability to generalise and operate comparably beyond data in their training distribution. We address this gap with DeepKnowledge, a systematic testing methodology for DNN-based systems founded on the theory of knowledge generalisation, which aims to enhance DNN robustness and reduce the residual risk of ‘black box’ models. Conforming to this theory, DeepKnowledge posits that core computational DNN units, termed Transfer Knowledge neurons, can generalise under domain shift. DeepKnowledge provides an objective confidence measurement on testing activities of DNN given data distribution shifts and uses this information to instrument a generalisation-informed test adequacy criterion to check the transfer knowledge capacity of a test set. Our empirical evaluation of several DNNs, across multiple datasets and state-of-the-art adversarial generation techniques demonstrates the usefulness and effectiveness of DeepKnowledge and its ability to support the engineering of more dependable DNNs. We report improvements of up to 10 percentage points over state-of-the-art coverage criteria for detecting adversarial attacks on several benchmarks, including MNIST, SVHN, and CIFAR.
\end{abstract}



\maketitle

\section{Introduction}
\label{sec:intro}
%
Deep Neural Networks (DNNs) exhibited enormous progress in achieving and in some instances surpassing, human-level performance in many challenging tasks~\cite{SutskeverVQ2014,HintonDYD2012,goodfellow2017challenge}. This success led to extensive DNN adoption in a wide spectrum of safety- and security-critical applications ranging from drug discovery in healthcare~\cite{litjens2017survey} to flight control systems~\cite{julian2016policy} and autonomous driving~\cite{bojarski2016end}.

Notwithstanding the remarkable advances, DNN models still present considerable inconsistency in their performance~\cite{zhang2019machine}, proven to be unstable with respect to small perturbations in the input data.
  Several safety incidents (e.g., Tesla's Autopilot crash~\cite{banks2018driver}), 
  resulted in high uncertainty and general mistrust in employing DNNs to undertake sensitive tasks in applications where high dependability is non-negotiable. 
Industrial studies have shown that DNN performance degrades considerably when data from the operational environment shifts away from the distribution assumed during training~\cite{zhou2019metamorphic}, raising major concerns about the robustness of the deployed model when facing unexpected data domain shift and/or adversarial perturbations~\cite{xu2020adversarial}. Due to their black-box nature, testing DNNs and detecting incorrect behaviours via conventional testing approaches is inadequate to ensure high DNN trustworthiness~\cite{BurtonGH2017}.

Stemming from traditional software testing, systematic DNN testing has emerged, aiming at understanding the capacity of DNNs to operate dependably during deployment when encountering corner input data not seen during the training phase~\cite{riccio2020testing,handelman2019peering}. 
White-box DNN testing informed by adequacy criteria has become a reference methodology because of its significant advances over ad-hoc testing and its effectiveness in detecting the diversity of test sets~\cite{pei2017deepxplore}. 
Illustrative examples from this line of research include testing techniques performing coarse- and fine-grained analysis at neuron level~\cite{pei2017deepxplore,ma2018deepgauge}, measuring the relative novelty of the input~\cite{sun2018testing} or leveraging explainability concepts~\cite{gerasimou2020importance} (see Section~\ref{sec:relatedwork}).
 
Despite the innovations, existing data-centric approaches do not necessarily help explain how the input features, computational units (neurons), and optimization algorithms of a DNN jointly contribute to achieving its strong generalisation power. 
Generalisation is crucial in a DNN, denoting its capability to correctly recognize unseen input data~\cite{wang2018identifying}, with generalization bound signifying the DNN's predictive performance~\cite{Reid2010}.
From a safety assurance perspective, assessing the capability of DNNs to respond appropriately when unexpected inputs and/ or corner cases appear in real-world scenarios entails establishing a functional understanding of the DNN components that learn to generalise~\cite{BurtonGH2017}. 
These DNN `learners' are core computational units that can generalise knowledge abstracted during training and apply it to a new domain without re-training.
These learners should be tested extensively to establish the capacity of the DNN to operate dependably upon deployment.

Driven by this insight, we introduce \approach, a systematic testing methodology accompanied by a knowledge-driven test adequacy criterion for DNN systems based on the out-of-distribution generalisation principle~\cite{Reid2010}. 
The hypothesis underpinning our approach is that by studying the generalisation capability of a DNN, we deepen our understanding of the model decision making process. 
\approach\ analyses the generalisation behaviour of the DNN model at the neuron level to capture neurons that jointly contribute to achieving the strong generalization power of the model not only within the training distribution but also under a domain (i.e., data distribution) shift.
Therefore, 
we employ ZeroShot learning~\cite{long2017zero} to estimate the model's generalisation capabilities under a new domain distribution. ZeroShot learning enables the DNN model to make predictions for classes that were not part of its training dataset.
By analysing each neuron's ability to abstract knowledge from training inputs and generalising it to new domain features, we  
establish a causal relationship between the neurons and the DNN model's overall predictive performance and identify \emph{transfer knowledge (TK) neurons}. 
The effective capacity of these transfer knowledge neurons for learning (reuse--transfer features from training to a new domain) positively influences the DNN generalisation behaviour and can explain which high-level features influence its decision-making. 
Consequently, these neurons are more critical for the correct DNN behaviour and should be allocated more testing budget. 
The TK-based adequacy criterion instrumented by \approach\ quantifies the adequacy of an input set as the ratio of combinations of transfer knowledge neuron neurons clusters covered by the set. We conduct a large-scale evaluation with publicly available datasets (SVHN~\cite{netzer2011reading}, GTSRB~\cite{Houben-IJCNN-2013}, CIFAR-10 and CIFAR-100~\cite{krizhevsky2010convolutional}, MNIST~\cite{deng2012mnist}) and multiple DNN models for image recognition tasks, demonstrating that
our approach can develop a functional understanding of the DNN generalisation capability and effectively evaluate the testing adequacy of a test set. Moreover, the findings show a substantial correlation between \approach's test adequacy criterion and the diversity and capacity of a test suite to reveal DNN defects as indicated by the coverage difference between the original test set and adversarial data inputs (cf. Section~\ref{sec:result}). 
Our main contributions are:
\vspace{-1.3mm}
\begin{itemize}
	\item The \approach\ approach for identifying \emph{transfer knowledge neurons} that have an influence on the DNN model's decision confidence under domain shift (Section~\ref{step2});
	\item The \approach\ coverage criterion which can establish the adequacy of an input set to trigger the DNN's knowledge generalisation behaviour, thus enabling to assess the semantic adequacy of a test set (Section~\ref{step3});
	\item An extensive experimental evaluation using state-of-the-art  DNN models, publicly-available datasets and recent test adequacy criteria (Section~\ref{sec:experiments});  and
	\item A prototype open-source \approach\ tool and repository of case studies, both of which are publicly available from our project webpage at \url{https://anonymous.4open.science/r/DeepKnowledge-1FBC} (Section~\ref{sec:implementation}).
\end{itemize}


Paper Structure.
Section~\ref{sec:background} overviews DNNs and DNN testing. 
Sections~\ref{sec:approach} and~\ref{sec:implementation} introduce \approach\ and its implementation. 
Section~\ref{sec:experiments} describes the evaluation of \approach.
Sections~\ref{sec:relatedwork}  and ~\ref{sec:conclusions} discuss related work and conclude the paper, respectively.

\section{Background \label{sec:background}}

\subsection{Deep Neural Network-based Systems}
\begin{figure}[t]
	\centering
	\includegraphics[width=0.6\linewidth]{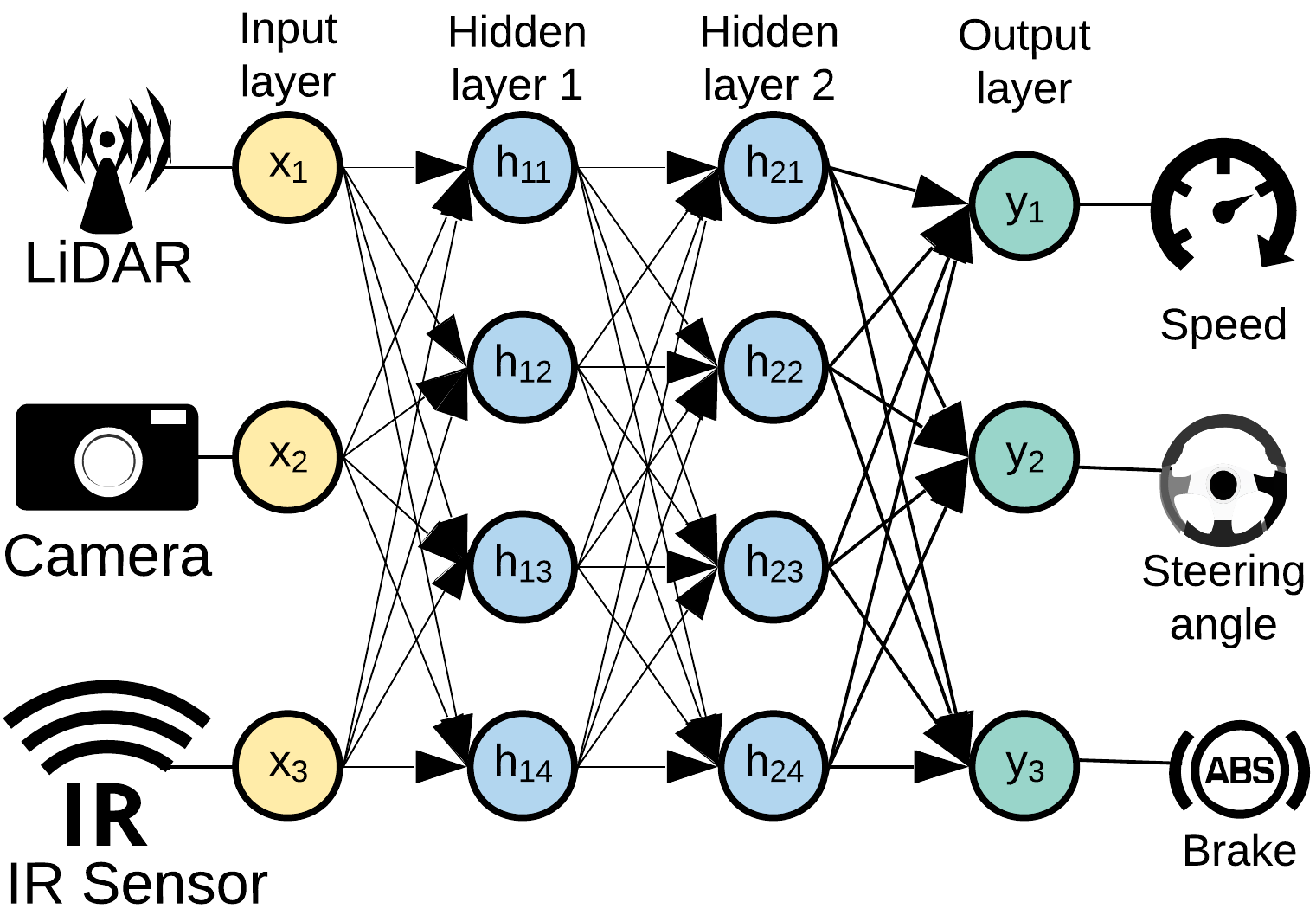}
	\caption{A four-layer fully-connected DNN system that receives inputs from vehicle sensors (camera, LiDAR, infrared) and outputs a decision for speed, 
		steering angle and brake.}
	\label{fig:dnn}
 \vspace{-6mm}
\end{figure}

\vspace{-1mm}
We consider any software system equipped with at least one DNN component as a DNN-based system. 
These systems benefited from the great achievements DNNs made 
across several application domains~\cite{LecunBH2015}. 
Unlike traditional machine learning algorithms, DNNs do not require intensive feature engineering~\cite{lenz2015deep}.
Inspired by neurobiology, the DNN architecture allows the network to extract features automatically from raw data without manual intervention or support from domain experts. 
A DNN uses mathematically complex computations to automatically extract high-level patterns from within the inputs to make a prediction~\cite{Goodfellow-et-al-2016}.
Several DNN architectures have been introduced, including 
GANs~\cite{goodfellow2020generative} and transformers~\cite{vaswani2017attention}. 
Overall, DNN can be classified into three main categories: multi-layer perceptrons (MLP)~\cite{kruse2022multi}, convolutional neural networks (CNN)~\cite{gu2018recent}, and recurrent neural networks (RNN)~\cite{mikolov2010recurrent}. 
As shown in Figure~\ref{fig:dnn}, independently of its type, a DNN comprises multiple interconnected neurons organized on hidden, input, and output layers. 
Neurons are computational units that combine multiple inputs and produce a single output by applying non-linear activation functions such as sigmoid, hyperbolic tangent and rectified linear unit (ReLU). 
The main phases of DNN are training (or learning), where the neural network abstracts knowledge (i.e., high-level features) from training data, and inference (or prediction) when the network applies the extracted knowledge to unforeseen inputs. 

\vspace{-1mm}
\subsection{Software Testing DNNs} 
\vspace{-1mm}
Software testing~\cite{pezze2008software} is a well-defined software engineering process, encompassing validation and verification, to establish whether the system under test meets a set of specified requirements.
More informally, it is an in-depth investigation of the system's behaviour to find bugs, errors, or missing requirements in the developed software.
A typical example is unit testing~\cite{runeson2006survey} 
that aims to tackle the smallest testable parts of an application, called units, and assess their correctness~\cite{hunt2004pragmatic,meszaros2007xunit}. 
An essential aspect of software testing is the test oracle~\cite{barr2014oracle}, which acts as a mechanism to assess whether the software executed a test case correctly. Conceptually, this involves comparing the oracle with the actual system output~\cite{pezze2008software}. 

To ensure the safety and correctness of DNNs, research focuses on adapting software testing methodologies such as test adequacy and code coverage to DNNs. Software coverage is typically quantified by counting the statements or paths executed by an input~\cite{pezze2008software}, aiming to maximize code coverage, as higher coverage implies more comprehensive testing.
However, following a purely traditional software testing methodology to DNN testing is not applicable due to: 
(i) the black-box nature of DNNs, which are data-driven and typically non-interpretable predictive models for which their decision-making process is unexplainable~\cite{montavon2017explaining}; and 
(ii) volume-related issue: DNN models are typically large structures whose parameters (number of layers and neurons per layer) can reach billions in size. 
Thus, it is inaccurate to compare neurons to software statements and layers to execution paths~\cite{zhang2019machine}. 
Instead, specialized testing criteria have been proposed for DNNs. These criteria are designed to consider the DNN structure and monitor the neurons and intrinsic network connectivity at various levels of granularity~\cite{zhang2019machine}.
We review related work in DNN testing and verification in Section~\ref{sec:relatedwork}.


\begin{figure}[t]
\centering
\centerline{\includegraphics[width=0.8\linewidth]
{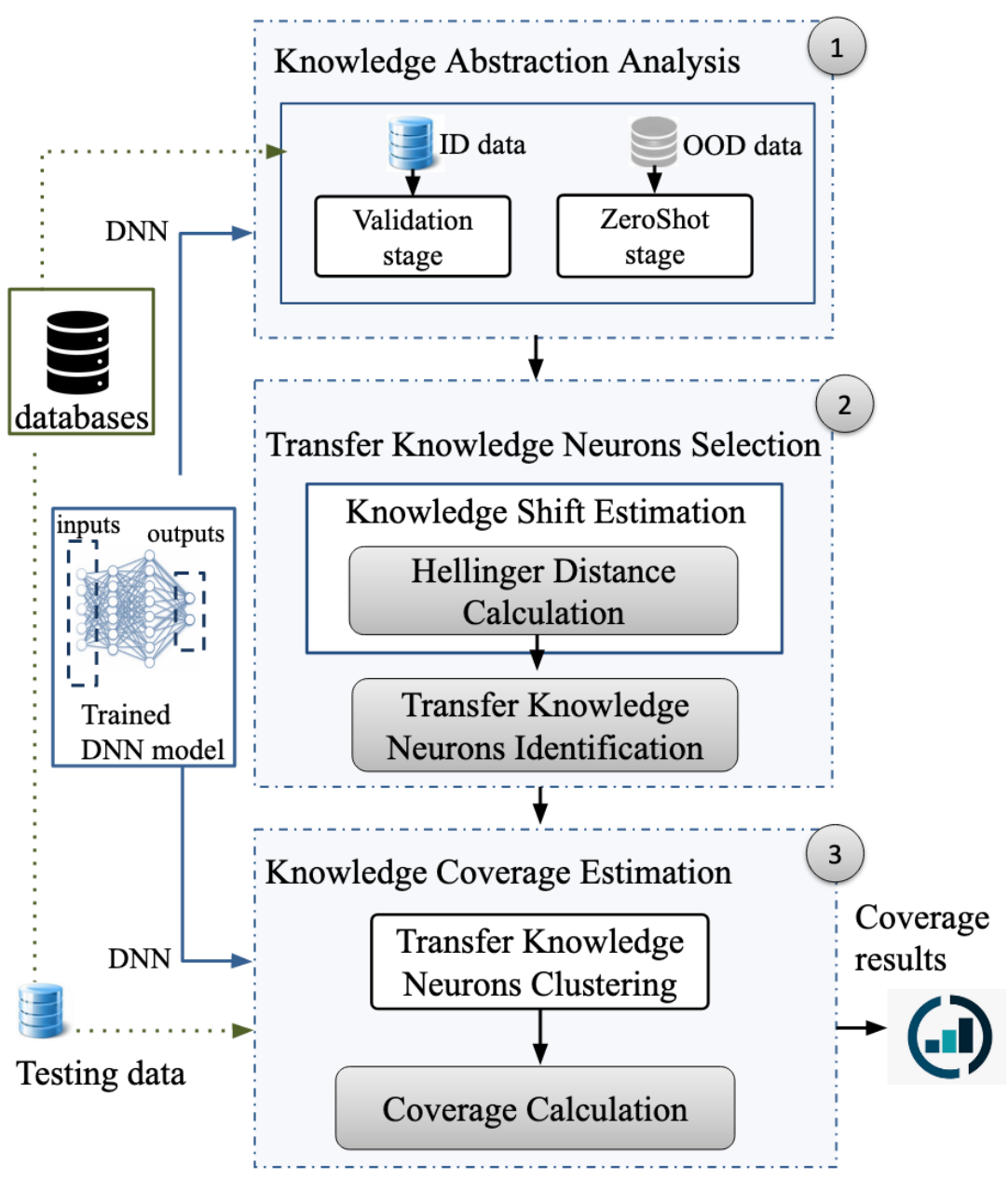}}
\caption{Overview of \approach}
\label{pipeline}
\vspace{-6mm}
\end{figure}

\section{\approach}
\label{sec:approach}

\approach, whose high-level workflow is shown in Figure~\ref{pipeline}, is a systematic testing approach for DNN systems. 
Using a trained DNN, an \textit{in-distribution} (ID) and an \textit{out-of-distribution} (OOD) dataset~\cite{yang2021generalized}, \approach\  analyses how the model learns and transfers knowledge abstractions under domain shift.
In this analysis, the learning problem is looked at through the prism 
of the DNN generalisation theory~\cite{urolagin2011generalization,kawaguchi2017generalization}, based on which \approach\ establishes a fundamental understanding of \textit{out-of-distribution generalisation} at the neuron level and quantifies the individual contribution of each neuron to this process. 
Through a filtering step, \approach\ identifies a set of \emph{Transfer Knowledge (TK) neurons}, which, given their contribution to the model's generalisation performance, are considered core DNN computational units.
The rationale behind using \textit{out-of-distribution generalisation} is that it enables the analysis of what part of the knowledge gained during DNN training can be employed in a new domain without re-training or fine-tuning. 
Thus, it allows the simulation of real-world corner cases the model could encounter during deployment. 
The next \approach\ step involves the execution of fine-grained clustering analysis to determine activation value clusters that reflect the changes in neurons' behaviour with respect to new inputs. 
The produced clusters of the transfer knowledge neurons are then used to  assess the coverage adequacy of the test set, conforming to the combinatorial analysis method defined in DeepImportance~\cite{gerasimou2020importance}. 

The selection of the OOD dataset is crucial in \approach.
The OOD data is regarded as a shift from the ID data, which by definition is the training dataset in DNN engineering~\cite{hendrycks2021many}. 
The distribution shift can be either: 
(i) a synthetic distribution shift resulting from a computer-generated perturbation, i.e., noise, adversarial attacks~\cite{papernot2016limitations}; or 
(ii) a natural distribution shift due to unseen input data~\cite{kurakin2016adversarial}. 
Since data distribution is critical in DNN testing, especially for the generalisation behaviour, in \approach, we carefully craft the OOD data based on: 
(i) the semantic similarity with the ID dataset; and 
(ii) the novelty compared to ID data samples. 
For instance, for a person detection DNN model trained on data of pedestrians, images of people cycling would be ID samples. In contrast, images of people with occlusion (i.e., hidden or `occluded' by real-world objects such as a car) would be OOD samples.

\subsection{\textbf{Knowledge Abstraction Analysis}\label{sec:knowledgeTransfer}}
Identifying the \textit{Transfer Knowledge (TK) neurons} within the DNN trainable layers is a key principle of \approach. 
Our approach begins by evaluating the prominence of each neuron and its connections to the OOD generalisation process~\cite{mccoy2019berts}. 
This analysis aims to identify neurons that can generalise knowledge abstracted during training and apply it to an OOD domain without re-training or fine-tuning. 
These TK neurons are core components of the DNN, collectively contributing to its generalisation behaviour, and, consequently, have a positive effect on the DNN accuracy and robustness~\cite{kawaguchi2017generalization}, enabling them to reach their peak performance.

As shown in Figure~\ref{pipeline}, for this analysis, we employ two different datasets, i.e., training samples and ZeroShot samples that correspond to \textit{in-distribution} and \textit{out-of-distribution} data, respectively. 
These datasets are used to quantify feature preference distributions over input features (e.g., image pixels for CNN models) for both the training and ZeroShot stages. 
This represents the knowledge abstraction analysis step that allows quantifying/identifying how the DNN model computational units, i.e., neurons, abstract/learn knowledge from the input feature space.
 
The identification of TK neurons is loosely inspired by recent research on quantifying knowledge change in DNN under domain shift at the level of individual neurons~\cite{rethmeier2020tx}. 
To this end, we employ the \textit{Activation maximization} explainability method~\cite{erhan2009visualizing} to derive a representation for features that neurons in the neural network have learned during the training and ZeroShot stages.
This method allows the identification of input instances that produce the highest activation values for a neuron, which are called \emph{`preferred'} input features.
Then, we turn these representations into probability distributions to quantify the shift in abstracted knowledge by utilizing a statistical metric on the defined space of probability distributions at the individual level (i.e., per neuron). Neurons able to achieve a certain threshold (that is empirically defined) of the statistical measure are selected as candidate transfer knowledge neurons.
 We leverage these insights and present the key contribution of this paper, namely \approach\ test adequacy criterion for testing DNN models by assessing test set quality and semantic diversity.


Now, let's formally explain this process.
Given a trained DNN model $D$ with $L$ trainable layers, each layer $L_i, 1 \leq i \leq L$ has $\left|L_i\right|$ neurons and the total number of neurons in $D$ is $S = \sum_{i=1}^{L} \left|L_i\right|$.
Let also $n_{ij}$ be the $j$-th neuron in the $i$-th layer. 
When the context is clear, we use $n \in D$ to denote any neuron that is a member of $D$ irrespective of its layer. 
Let $X$ denote the input domain of $D$, representing either the in-distribution (ID) or out-of-distribution (OOD) dataset,  and $x_k \in X$ be the $k$-th concrete input from $X$ (also termed input feature). Finally, we use the function $\phi(x_k,n_{ij}) \in \mathbb{R}$ to signify the output of the activation function of neuron $n_{ij} \in D$.

Given the input set $X$, $D(x_k) = \left(\phi\left(x_k,n_{11}\right), \ldots, \phi\left(x_k,n_{L_L|L_L|}\right)\right)$ enables calculating the activation trace of input $x_k \in X$. 
Using this information, we signify with $\overline{n^{x_k}} = argmax_{n \in D}\left(D(x_k)\right)$ the maximally activated neuron and with $\overline{\alpha^{x_k}} = max\left(D(x_k)\right)$ the corresponding maximum activation value. 
Applying these steps to the entire set $X$, we can construct the matrix
$M = \left[\left(x_k, \overline{n^{x_k}}, \overline{\alpha^{x_k}}\right) \right]_{\forall x_k \in X}$, 
where the $r$-th entry of $M$ 
captures the maximally activated neuron $\overline{n^{x_{k}}}$ and its activation value $\overline{\alpha^{x_{k}}}$ for input $x_k$.
Using methods like Monte Carlo Dropout~\cite{srivastava2014dropout,gal2016dropout} can yield matrix entries with different maximally activated neurons and/or values for the same input $x_k$.

Next, \approach\ establishes the 
input probability distribution per neuron~\cite{rethmeier2020tx} 
by aggregating the maximum activation values $\overline{\alpha^{x_k}}$ for each neuron $n$ over its maximally activated input features $x_k$. 
Accordingly, $A(n_{ij}) = \{ \left(x_k,\mu_{x_k}\right) \;|\; \forall x_k \in X : M_{n_{ij}}=\bigcup_{1\leq r \leq \left|M\right|} (x_k, n_{ij}, \overline{\alpha^{x_{k}}}) \;\bullet\; \mu_{x_k} = \frac{1}{\left|M_{n_{ij}}\right|} \sum_{\left(x,n,\alpha\right)\in M_{n_{ij}}}\alpha\}$. 
More informally, $A(n_{ij})$ provides the discrete input activation distribution per neuron where $\mu_{x_k}$ is the mean maximum activation of input $x_k$ for the $n_{ij}$ neuron. 
We can then use the set $A(n_{ij})$ to calculate each neuron's normalised probability distribution per $x_k$ as:
\begin{equation}
\label{eqP}
P_{n_{ij}}=\left\{ \left(x_{k},p_{k} \right )\right\}_{x_k \in X}, 
\qquad\quad p_{k}=\frac{\mu_{x_{k}}}
{\sum_{(x_l, \mu_{x_{l}}) \in A(n_{ij})}\mu_{x_{l}}}
\vspace{-1mm}
\end{equation}
where $p_{k}$ is the activation probability of input feature $x_{k}$  
given as the mean activation $\mu_{x_k}$ over the sum of its activation means. 


Finally, collecting the $P_{n_{ij}}$ enables assembling the DNN's activation distributions defined as $P={P_{n_{ij}},\cdots ,P_{n_{L_L\left|L_L\right|}}}$ over the dataset $X$. 
Doing this both for the ID and OOD datasets enables assembling the corresponding $P^{ID}$ and $P^{OOD}$ elements. 


Algorithm~\ref{knwTrQuan} shows the high-level process \approach\ employs for quantifying knowledge change in model $D$. 
Starting with the empty matrix $M$ and activation distribution $P$ (line 2), for any input $x_k \in X$, we perform standard forward passes to compute the activation trace and identify the maximally activated neuron $\overline{n^{x_{k,r}}}$ and the corresponding maximum activation value $\overline{\alpha^{x_{k}}}$ (lines 3 -- 8).  
Then, the matrix $M$ of neurons input max activations is updated (line 7), which we aggregate later to express each neuron as a probability distribution over maximally activated inputs as defined in equation \eqref{eqP} 
(lines 10-17).
Given the input set $X$, each neuron $n$ is expressed as a distribution over preferred inputs (line 16), which is returned when the algorithm terminates (line 19).
Executing Algorithm~\ref{knwTrQuan} for both $X^{ID}$ and $X^{OOD}$ datasets, yields the distribution over preferred inputs $P^{ID}$ and $P^{OOD}$, respectively. 

\begin{figure}[t]
\vspace{-2mm}
\begin{algorithm}[H]
\caption{Abstracted Knowledge Quantification}
\label{knwTrQuan}
\begin{algorithmic}[1]
  
    \Function{AbstractedKnowledgeQuantification}{$D$, $X$}
    \State $M \gets \emptyset$, $P \gets \emptyset$

    
    \For{$x_{k} \in X$} \Comment{$X$ $\in$  $\{X^{ID}, X^{OOD}\}$}
        \State $\overline{n^{x_k}}=argmax(D(x_{k}))$ 
        \State $\overline{\alpha^{x_k}}=max(D(x_{k}))$
        \State $v \gets \left(x^{k},\overline{n^{x_k}},\overline{\alpha^{x_k}} \right)$ \Comment{input's activation row vector}
        \State $M = M \cup  v $
        \Comment{neuron \textit{preferred} inputs matrix}
    \EndFor
    \For { $n \in D$} \Comment{using $n$ (not $n_{ij}$) for simplification}
        \State $A(n) \gets \emptyset$
        \For{$x_{k} \in X$}
            \State $M_{n_{ij}} = \textsc{ExtractFromM}(x_k, n)$ 
            \State $\mu_{x_k} = \frac{1}{\left|M_{n_{ij}}\right|} \sum_{\left(x,n,\alpha\right)\in M_{n_{ij}}}\alpha$
            \State $A(n) = A(n) \cup \left(x_k, \mu_{x_k}\right)$
        \EndFor
        \State $P_{n}=\left\{ \left(x_{k},p_{k} \right )\right\}_{x_k \in X}$ \Comment{using \eqref{eqP}}
        
        \State $P = P \cup P_n$
    \EndFor 
    \State \textbf{return} $P$ \Comment{$P^{ID}$ for $X^{ID}$, $P^{OOD}$ for $X^{OOD}$}
    \EndFunction

\end{algorithmic}
\end{algorithm}
\vspace{-12mm}
\end{figure}


\subsection{\textbf{Transfer Knowledge Neurons Selection}}
\label{step2}
Next, \approach\ determines the \emph{transfer knowledge (TK) neurons} by quantifying the \emph{change} and \emph{diversity} in the knowledge abstracted per neuron $n_{ij}$ from the ID dataset to the OOD dataset using the activation distribution probabilities $P^{ID}$ and $P^{OOD}$.
Our approach estimates \emph{knowledge change} by employing statistical difference measures between the preference features distributions calculated per dataset. 
Statistical distance measures are commonly used to measure distribution shifts~\cite{hendrycks2021many}. 
\approach\ leverages the \textit{Hellinger Distance} (HD), a symmetric divergence measure that quantifies the deviation between two distributions, i.e., it measures how far apart the two \textit{features preferred distributions} are in the space of probability distributions common to them.
%
Concretely, \approach\ uses HD to measure the per-neuron knowledge change between the activation distribution probabilities over the ID and OOD datasets, i.e., $p=P_{n_{ij}}^{ID}$ and $q=P_{n_{ij}}^{OOD}$ given by
\vspace{-1mm}
\begin{equation}
\label{HD}
HD(p,q)= \frac{1}{\sqrt{2}}\sqrt{\sum_{j=1}^{X}(\sqrt{p_{j}}-\sqrt{q_{j}})^{2}} \;\in [0,1]
\vspace{-2mm}
\end{equation}
HD is a bounded metric, taking values between 0 and 1; when $HD\!=\!0$, the two distributions are identical, and when $HD\!=\!1$, they are the furthest apart.
Thus, a high distance  for the $ij$-th neuron's \textit{features preferred distributions} under domain shift ($P_{n_{ij}}^{ID}, P_{n_{ij}}^{OOD}$), illustrates that the neuron did not generalize its abstracted knowledge to the new domain features well. 
The lower the HD value, the more the neuron can transfer learned knowledge to the OOD data space.
An empty distribution over $P_{n_{ij}}^{ID}=0$ or $P_{n_{ij}}^{OOD}=0$ results in an invalid HD value, which is assigned the smallest value possible (0). 

Note that \approach\ can employ other divergence metrics, including Wasserstein distance~\cite{vallender1974calculation} 
and Jensen–Shannon~divergence~\cite{menendez1997jensen}. 
We emphasise the importance of using symmetric distance measures  as they have the advantage of providing a unique interpretation of the knowledge shift, i.e., $distance(P_{n_{ij}}^{ID} || P_{n_{ij}}^{OOD})= distance(P_{n_{ij}}^{OOD}|| P_{n_{ij}}^{ID})$, while also 
requiring less computation time than 
asymmetric divergence measures such as Kullback-Leiber~\cite{joyce2011kullback}.

In addition to the \emph{knowledge change} given by the HD value per neuron, \approach\ uses \emph{knowledge diversity} between the ID and OOD datasets to extract the TK neurons. 
Similarly to~\cite{rethmeier2020tx}, \approach\ leverages the \textit{feature length} $l$  as a knowledge diversity indicator. 
To this end, for each neuron $n_{ij}$, we quantify the total number of unique preferred inputs denoted by $l(\overline{X_{n_{ij}}})$, where
$\overline{X_{n_{ij}}} = \{ x_k \;|\; \forall \left(x_k,p_k\right) \in {P}_{n_{ij}} \bullet p_k > 0\}$. 
As before, the set and its length can be constructed both for the ID and OOD datasets. 
The rationale for this knowledge diversity analysis is that input features that have maximally activated the same neuron in the ID and OOD datasets can be considered semantically similar and can be deployed to quantify the generalization capabilities of the neuron.
The analysis enables further the elimination of neurons whose HD value is low but do not have many preferred inputs, and thus their $l(\overline{X_{n_{ij}}})$ is small either for the ID or OOD datasets.

We define three neuron types based on $l(\overline{X_{n_{ij}}})$~\cite{rethmeier2020tx}:
(1) \textit{`Gained'} neurons, where $l(\overline{X^{ID}_{n_{ij}}})<l(\overline{X^{OOD}_{n_{ij}}})$, indicating that these neurons have gained knowledge in ZeroShot learning and can specialize in the new domain; 
(2)  \textit{Avoided} neurons, where $l(\overline{X^{ID}_{n_{ij}}})>l(\overline{X^{OOD}_{n_{ij}}})$, meaning that they lost part of the learned knowledge during the ZeroShot stage; and
(3) \textit{`Stable'} neurons, where $l(\overline{X^{ID}_{n_{ij}}})=l(\overline{X^{OOD}_{n_{ij}}})$.

Employing the \emph{knowledge change} and \emph{knowledge diversity} mechanisms presented above, \approach\ selects the set of transfer knowledge TK neurons. 
The HD threshold  and the neuron type (gained, avoided, preferred) are hyperparameters of \approach\ for knowledge change and diversity estimation, respectively. 
We illustrate in our empirical evaluation (Section~\ref{sec:experiments}) what constitutes appropriate parameters
($HD \in (0.01-0.05]$ and gained).



\subsection{\textbf{Knowledge Coverage Estimation}}\label{step3}
To increase our confidence in the robustness of the DNN model's behaviour, we assess the test set adequacy, i.e.,
how well the test set exercises the set of identified TK neurons.
First, \approach\ determines regions within the TK activation value domain central to the DNN execution and clusters them into separate combinations.  
Each combination of clusters reflects different knowledge abstracted during the training phase.
We can assess the knowledge diversity of the test set by assessing if it adequately covers these combinations, similar to combinatorial interaction testing in conventional software testing~\cite{petke2015practical}.
We posit that semantically similar inputs, i.e., inputs with similar features, generate activation values with a similar statistical distribution at the neuron level~\cite {gerasimou2020importance,MJX18}.

More specifically, we adopt the combinatorial neuron coverage method introduced in~\cite{gerasimou2020importance} to first iteratively cluster the vector of activation values of each TK neuron from the training set (cf. Section~\ref{sec:knowledgeTransfer}). 
Then, given inputs from the testing set, we measure the number of combinations this set covers over the combinations of activation value clusters of the TK neurons. 
Since the cluster combinations of activation values generated correspond to semantically different features of the input set, this combinatorial analysis could assess the knowledge diversity of the test set. 

Formally, we first identify regions within the TK value domain that are key to the DNN, i.e., clusters of activation values using  iterative unsupervised learning via the combinations of k-means clustering~\cite{likas2003global} and the Silhouette index~\cite{rousseeuw1987silhouettes}.
Clustering enables dimensionality reduction of neuron activation values, with the Silhouette index employed to reinforce the clustering output.
Silhouette is an internal clustering validation index that computes the goodness of a clustering structure without external information, facilitating the selection of the optimal number of clusters. 
%
The Silhouette score for $c \in \mathbb{N}^{+}$ clusters for neuron $n \in D$ is given by:
\vspace{-2mm}
\begin{equation}
S_{n}^{c}=\frac{1}{\left | X \right |}\sum_{x=1}^{X}\frac{E\left ( x \right )- I\left ( x \right )}{max \left ( E\left ( x \right ),I\left (x \right ) \right )}
\end{equation}

\noindent
where $I\left ( x \right )$ is the intracluster cohesion, i.e.,  the average distance of activation value $\phi \left ( x,n \right )$ to all other values in the same cluster, and $E\left ( x \right )$ is the inter-cluster separation, i.e., the distance between $\phi \left ( x,n \right )$ and activation values in its nearest neighbour cluster.

Next, we measure the degree to which a test set $X$ covers the clusters of TK neurons, by evaluating the combinations of activation value clusters triggered by inputs $x \in X$.
Thus, having the set of 
clusters extracted and optimized, we identify the vector of TK neurons cluster combinations (TCC) as follows:
\vspace{-1mm}
\begin{equation}
\label{eq::TCC}
TCC=\prod_{n \;\in\; TK}\left \{ Centroid\left ( \Psi _{n}^{i}|\: \forall 1\leq i \leq \left | \Psi _{n} \right | \right ) \right \}
\end{equation} 
\noindent
where the set $\Psi_n^i$ contains the activation values of cluster $i$ for the $n$-th TK neuron and $Centroid(\Psi_{i}^{n})$ gives its centre of mass.
For more details on cluster extraction and $\Psi_{n}^{i}$ estimation, please see~\cite{gerasimou2020importance}. 

Finally, given the input set $X$, the \approach\ \emph{Transfer Knowledge Coverage (TKC)} score is computed as the ratio of TCC covered by all $x\in X$ over the size of the TCC set as follows:
\begin{equation}
\mathrm{TKC}=\frac{|\{  TCC{\left ( i \right )}|\exists  x \in X : \forall V_{n}^{j} \in  TCC{\left( i \right)} \bullet min \; d( \phi ( x,n), V_{n}^{j})\}|}{\left| TCC \right|}
\end{equation}
TCC{(i)} is covered if there is an input $x$ for which the Euclidean distance $d(\phi(x,n), V_{n}^{j})$ between the activation values of all TK neurons and the $i$-th neuron's cluster centroids is minimised. 
Overall, the TKC score quantifies if the combinations of activation value clusters of TK neurons have been covered adequately by the test set. 
A higher TKC value entails a knowledge-diverse test set executing different TK neuron cluster combinations, helping to assess the test set's knowledge diversity.
TKC is sound by construction; adding another input $x'$ in $X$ will either preserve or increase the TKC score.

\section{Implementation and hardware\label{sec:implementation}}
All experiments were conducted on a high-performance computer running a cluster GPU with NVIDIA 510.39. We implement \approach\  using the state-of-the-art framework, open-source machine learning framework Keras (v2.2.2)~\cite{gulli2017deep} with Tensorflow (v2.6) backend. 
We report the full experimental results next.
To allow for reproducibility, our open-source implementation and evaluation subjects are available on \approach's Github repository at \url{https://anonymous.4open.science/r/DeepKnowledge-1FBC}.

\section{Experimental Evaluation\label{sec:experiments}}
\subsection{Research Questions\label{sec:rqs}}
\textbf{RQ1  (Knowledge Generalisation)}: 
Is \approach\ capable of capturing the most influential neurons of a DNN system?

\vspace{0.5mm}\noindent
\textbf{RQ2  (Hyperparameter Sensitivity)}: 
How do \approach's hyperparameters that encode knowledge change and knowledge diversity affect the execution of the approach?


\vspace{0.5mm}\noindent
\textbf{RQ3 (Effectiveness)}: 
Can the TKC criterion of \approach\ be effectively used as a test adequacy criterion?

\vspace{0.5mm}\noindent
\textbf{RQ4 (Correlation)}: 
How does  TKC  perform against state-of-the-art  DNN testing criteria~\cite{pei2017deepxplore,ma2018deepgauge,gerasimou2020importance,kim2019guiding}?


\vspace{0.5mm}\noindent
\textbf{RQ5  (OOD Selection )}: What is the impact of the out-of-distribution dataset on the selection of TK neurons and the coverage results? 

\subsection{Experimental Setup}
\textbf{Datasets and DNN-based Systems.} 
We extensively evaluate \approach\ on four different DNN-based systems using: (1) the  original  test  sets; and (2)  adversarial  examples  generated  by four  attack  strategies.
Table~\ref{table:Dl_data} lists the widely-adopted datasets employed to evaluate \approach. 
In particular, we used
MNIST (28x28 grayscale images of handwritten digits), SVHN (32x32 coloured images of real-world house numbers), 
and CIFAR-10 (32x32 coloured images from 10 classes) as in-distribution (ID) data, and as out-of-distribution datasets (OOD): Fashion MNIST (a dataset of Zalando's article -- 28x28 grayscale images), EMNIST (MNIST extension with both handwritten digits and letters -- 28x28 pixel images), CIFAR-100 (dataset of 32x32 colour images from 100 different classes). 
The GTSRB dataset~\cite{Houben-IJCNN-2013} (from the German Traffic Sign Recognition Benchmark) with over 50,000 images classified into 40 classes was also used for RQ5.
We should highlight that the selected OOD dataset should share features with the ID dataset (e.g., image size, colour)~\cite{hendrycks2021many}. 
\noindent
We performed a comprehensive assessment of \approach\ using DNN-based systems from recent research~\cite{zhang2020testing}. 
The majority of employed DNN models are convolutional DNNs with different architectures, trained with different datasets, making them suitable for the empirical analysis of \approach\ and TKC as an adequacy criterion.
We study LeNet-1 and  LeNet-5 from the LeNet family~\cite{lecun1998mnist}, and the ResNet18 model~\cite{targ2016resnet}.

\begin{table}[t]
	\centering
	{\small
	    \renewcommand{\arraystretch}{1.05}
		\caption{Datasets and DNN models used in our experiments}
        \label{table:Dl_data}
		\vspace*{-1mm}
		\begin{tabular}{p{1.6cm}p{2.4cm}p{2.2cm}p{1cm}}
			\hline
			\textbf{$\!\!\!$ID Dataset} 
			    & $\!\!\!$\textbf{OOD Dataset}
			    & $\!\!\!$\textbf{DNN Model}
                & $\!\!\!$\textbf{\# Layers}
            \\
			\hline
        MNIST~\cite{deng2012mnist}    
            &Fashion MNIST~\cite{xiao2017fashion}   
            & LeNet1 \& 4~\cite{lecun1998mnist}   
            & 5
        \\
        SVHN~\cite{netzer2011reading}     
            & EMNIST~\cite{cohen2017emnist}
            & LeNet5~\cite{lecun1998mnist}   
            & 7
        \\
        CIFAR-10~\cite{krizhevsky2010convolutional} 
            & CIFAR-100~\cite{krizhevsky2009learning}  
            & ResNet18~\cite{zhang2019pedestrian} & 18 
        \\\hline
		\end{tabular}
	}
    \vspace{-4mm}
\end{table}


\vspace{1mm}\noindent
\textbf{Adversarial Examples.}
We use the following techniques to evaluate \approach\ through adversarial examples: Fast Gradient Sign Method (FGSM)~\cite{liu2019sensitivity}, Basic Iterative Method (BIM)~\cite{goodfellow2014explaining}, Momentum Iterative Method (MIM)~\cite{dong2018boosting}, and Projected Gradient Descent (PGD)~\cite{deng2020universal}. 
Adversarial inputs are crafted using the Cleverhans Python library~\cite{papernot2016cleverhans} to generate perturbed examples based on given benign inputs.  
For each of the original test sets (SVHN, MNIST and CIFAR-10), we generate an adversarial set of equal size.

\vspace{1mm}\noindent
\textbf{Coverage Criteria Configurations.}
We used the following \approach\ parameter ranges in the experiments:
$HD \in [0.01, 0.25]$, $|TK| \in \left \{20\%, 30\%,40\%\right \}$ of the total candidate TK neurons and all three knowledge diversity types, i.e., Gained, Avoided, and Stable (Section~\ref{step2}).
When comparing \approach\ against state-of-the-art coverage criteria for DNN systems, for each criterion we use the hyperparameters recommended in its original research paper. 
To this end, for all research questions, we set the threshold for neuron coverage (NC)~\cite{pei2017deepxplore} to 0.7. For k-multisection neuron coverage (KMNC) and top-k neuron coverage (TKNC)~\cite{ma2018deepgauge}, the number of multisections $k=10$ and $k=3$, respectively.
For neuron boundary coverage (NBC) and strong neuron activation coverage (SNAC)~\cite{ma2018deepgauge}, we set as the lower/upper bound the minimum/maximum activation values encountered in the training set. 
For likelihood-based surprise coverage (LSC)~\cite{kim2019guiding}, we use the last convolutional layer, with 
the upper bound set to 2000 and using 1000 buckets.

\subsection{Results and Discussion}
\label{sec:result}\vspace{-1mm}
\noindent
\textbf{RQ1 (Knowledge Generalisation).} 

In this RQ, We experiment with our test coverage criterion on state-of-the-art DNNs that
have a broad range of sizes to demonstrate the utility of our criteria with respect to the Knowlwdge generalisation efficiency of testing input generation,
In this research question, we investigate
whether the selected TK neurons can be used to instrument the TKC test adequacy criterion. 
To answer this research question, we employ a \approach-based model retraining strategy where we augment the training set with a set of input images selected by the \approach. 
The input images that maximally activate the TK neurons are identified and used to retrain the DNN. If these inputs are effective in improving the model accuracy, then, \approach\ is able to select semantically meaningful data that improve the model generalisation capabilities.

\begin{table}[t]
\renewcommand{\arraystretch}{1.05}
\small
\centering
\caption{Effect of TKC-based retraining on DNN accuracy}
\label{tab:knwdata}
\vspace{-1mm}
\begin{tabular}{p{1.5cm}p{1.3cm}p{0.9cm}p{0.9cm}p{0.9cm}p{0.8cm}}
\hline
    \textbf{DNN}
        &\textbf{Dataset}  
        &\textbf{Initial} 
        &\textbf{$T2_{TKC}$} 
        &\textbf{$T2_{R}$} 
        &\textbf{$T2_{A}$} 
    \\\hline
    LeNet1                                      
        & MNIST    
        & 97.75\%                   
        & 98.84\%       
        & 91.69\%          
        & 9.74\%       
    \\\hline
    \multicolumn{1}{l}{\multirow{3}{*}{LeNet5}} 
        & MNIST    
        & 98.10\%                   
        & 99.04\%       
        & 98.89\%          
        & 9.32\%        
        \\ 
    \multicolumn{1}{c}{}                        
        & SVHN     
        & 84.51\%                  
        & 85.12\%       
        & 84.18\%          
        & 19.58\%       
        \\ 
    \multicolumn{1}{c}{}                        
        & CIFAR-10  
        & 93.05\%                   
        & 99.88\%       
        & 89.94\%           
        & 10.97\%        
        \\ \hline
    ResNet18                                    
        & SVHN     
        & 95.77\%                   
        & 95.90\%       
        & 95.56\%          
        & 6.69\%         
        \\ \hline

\end{tabular}
\begin{tablenotes}
\vspace{-0.5mm}
\small
      \item \textcolor{darkgray}{*Initial, $T2_{TKC}$, $T2_{R}$ and $T2_{A}$ denote the  initial DNN accuracy, and after \approach-, Random- and Adversarial-based retraining, respectively.} 
\end{tablenotes}
\vspace{-6mm}
\end{table}

We split our ID test set into separate subsets $T_{1}$ and $T_{2}$ using a 60\% - 40\% ratio. 
Then, we identified input images that improve \approach\ coverage (i.e., maximally activated the TK neurons) from within the $T_{2}$ subset, 
producing \approach-guided input sets $T2_{TKC}$. 
We highlight that only inputs that produce the highest activation for all TK neurons are kept. 
This resulted in augmented sets limited in size, i.e., 6K, 3K, and 5K, for MNIST, SVHN, and CIFAR-10, respectively.
Similarly, we generated augmented sets $T2_{R}$ 
using randomly selected neurons, ensuring that none of the $T2_{R}$ 
inputs is included in $T2_{TKC}$. 
The third augmented set  $T2_{A}$ 
is generated using the PGD adversarial technique. 
Finally, we retrain the DNN models separately using the three different augmented datasets: $T2_{TKC}$, $T2_{R}$ and $T2_{A}$.
Based on empirical observations that DNN retraining  on small datasets shows a performance decline after the 10th epoch, we stop retraining after that point.
    		
Table~\ref{tab:knwdata} reports the DNN prediction accuracy on the $T_{1}$ testing sets  after retraining. 
For each model, we report the accuracy of the original model without retraining (`Initial'), and the prediction accuracy after the \approach-based ($T2_{TKC}$), random-based ($T2_{R}$) and PGD-based ($T2_{A}$) retraining strategies.
The accuracy of each model after retraining with a separate test set shows that \approach-based augmentation strategy using $T2_{TKC}$ achieved on average $1\%$ higher accuracy improvement over the initial accuracy, e.g., from $97.75\%$  to $98.84\%$  for LeNet1-MNIST and from $98.10\%$  to $99.04\%$ for LeNet5-MNIST. 
Unsurprisingly, the adversarial-based augmentation strategy ($T2_{A}$) produced a set that caused a significant drop in accuracy, from $97.75\%$ to $9.74\%$ for LeNet1-MNIST. 
Clearly, the limited size of the $T2_{TKC}$, e.g., only 5K instances have been generated for the CIFAR-10 dataset, contributes to the small rise in the DNN accuracy after \approach-based retraining.
Although the improvement made by $T2_{TKC}$ was limited (in some cases similar in comparison with the $T2_{R}$ set), the consistency in this behaviour over the 3 models and 3 different datasets lead to the conclusion that \approach-based augmentation strategy is efficient in generating data instances that improve the DNN accuracy. 
The $T2_{TKC}$ set improved the predictive accuracy and general performance of the different tested models. It added new variances of data through samples that had a direct impact on the learning process. 
Thus, these inputs are semantically significant to the generalisation process of a DNN. 
These observations provide insights into the sensitivity of TK neurons 
to semantically relevant input features, which entails that TK neurons are key contributors to the model generalization capability. 

Based on these findings, we conclude that \textbf{\approach\ can identify neurons that significantly contribute to the generalization process and, as a result, influence the DNN decision-making process. 
Thus, the TK neurons can be efficiently utilized as a test adequacy criterion to evaluate the testing adequacy of a test set.}
Further, \approach\ can be successfully deployed to adjust the DNN training phase by augmenting the training dataset to improve the predictive accuracy and general performance, thus reducing its erroneous behaviour. 

\vspace{2mm}\noindent
\textbf{RQ2 (Hyperparameter Sensitivity).}
We examined the impact of \approach's hyperparameters used for encoding knowledge change and knowledge diversity. 
To this end, we varied (i) the percentage of selected TK neurons, i.e., 
$|TK| \in \left \{20\%, 30\%,40\%\right \}$ of the total candidate TK neurons resulting from Step 2 of \approach; (ii) the HD threshold, i.e.,  $HD \in [0.01, 0.25]$; and (iii) the neuron knowledge diversity types, i.e., Gained 
and Stable. 
%
%
The results show that the TK set size varies between different DNNs. 
In some cases, this variation is extensive: 920 for LeNet1, 883 for LeNet5 and 1289 for ResNet18. 
The comparison of the HD bound for different DNN and dataset combinations showed that the $HD$ distribution of candidate neurons has an upper bound of $0.25 \pm 0.01$. 


An effective testing criterion should provide fine-grained insight into the DNN behaviour and also be hard, but not impossible, to cover~\cite{zhang2019machine}. 
Since \approach\ is underpinned by the knowledge generalisation principle~\cite{neyshabur2017exploring}, we aim at selecting a subset of the TK neurons that satisfy the main generalisation criterion of \emph{optimal capacity}, which provides a balance between overfitting and learning new knowledge.
To evaluate the validity of a candidate TK neuron $n$ against this criterion, we empirically analyse the following combinations of knowledge change 
$HD \in \{(0.01- 0.05], (0.10-0.15], (0.15-0.20], (0.20-0.25] \}$ 
and knowledge diversity: \textit{Gained}, $l(\overline{X^{ID}_{n}})<l(\overline{X^{OOD}_{n}})$, or \textit{Stable},  $l(\overline{X^{ID}_{n}})=l(\overline{X^{OOD}_{n}})$.
Then, given the set of candidate TK neurons, we select the top $p$ percentage with $p \in \left\{10\%,20\%,30\%,40\%,50\% \right\}$, forming the transfer knowledge neuron sets $TK_{10}, TK_{20}$ etc.

\begin{figure}[t]
	\centering
	\includegraphics[width=0.92\linewidth]{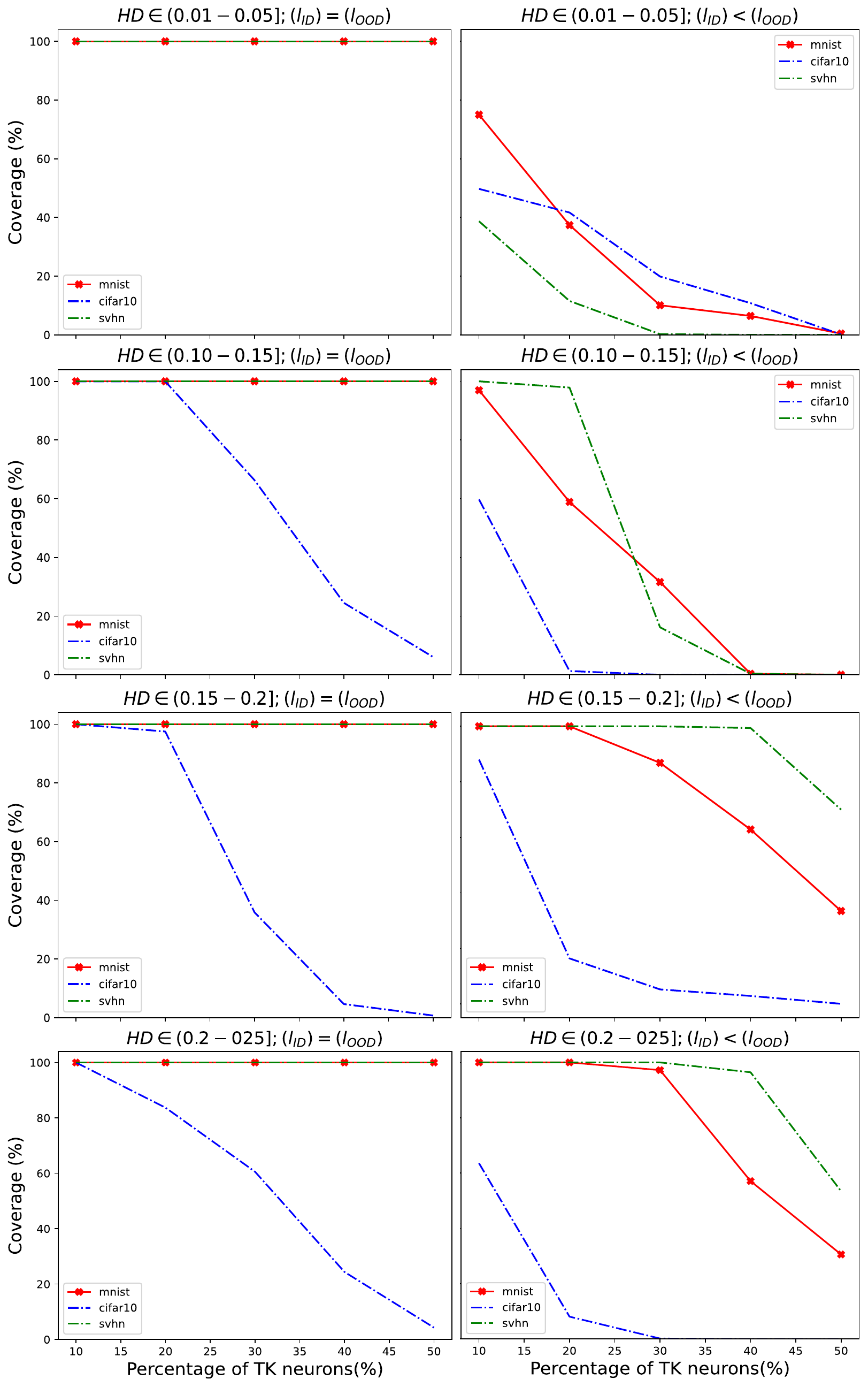}
    \vspace{-0mm}
    \caption{\approach\ coverage with different hyperparameters $HD$ and $l$ settings for different datasets.}
    \label{fig:vizparameters}
    \vspace{-6mm}
\end{figure}

As shown in Figure~\ref{fig:vizparameters}, overall the average \approach\ coverage values decrease when the analysis is performed on a bigger percentage of TK neurons (e.g., $TK_{40}$). 
By analyzing the effect of knowledge diversity (i.e., Gained vs Stable) on the average results of \approach, we observe Stable TK neurons perform poorly as a test adequacy criterion as the coverage becomes saturated and stagnates for most of the tested datasets (see the left column in Figure~\ref{fig:vizparameters}), i.e., the coverage reaches its peak already with only $TK_{10}$ of the TK candidate neurons set. 
For instance, for MNIST and SVHN, the \approach\ coverage value stagnated at 100\% even with the increase of the percentage of used TK neurons from $TK_{10}$ to $TK_{50}$. 
We observe the same behaviour irrespective of the selected HD values.
Accordingly, this can be interpreted as a failure of these neurons to generate diverse cluster combinations (TCC) and thus to abstract diverse semantic features from the dataset.

We notice that the selected TK neurons featured as Gained,
result in coverage values (right column in Figure~\ref{fig:vizparameters}) that moderately decrease as the percentage increases (from $TK_{10}$ to $TK_{50}$).
This criterion is also shown to be more difficult to cover by the test sets.
Moreover, \approach\ coverage with $HD \in (0.01-0.05]$ entails the ability to assign degraded coverage value according to the size of TK neurons set, with an adequacy criterion gradually harder to cover compared to the other neurons belonging to $HD$ ranges $(0.10-0.15]$, $(0.15-0.20]$ or $(0.20-0.25]$. 

The results show that Gained TK neurons with  $HD \in (0.01-0.05]$ are more difficult to cover, leading to a coverage metric that gradually decreases as the percentage $p$ increases.
The activation values of those TK neurons are concentrated in specific regions within their inputs domain. 
Informally, those regions capture features shared between ID and OOD datasets and are the most influential for learning. 
Thus, the combinations of TK neuron clusters are less likely to be covered if testing inputs do not introduce new semantically diverse features  (compared to those from the ID dataset).

Table~\ref{tab:par} illustrates these results in more detail, showing the average TKC value for different DNN models and percentages of Gained TK neurons with $HD \in (0.01-0.05]$. 
To reduce randomisation bias, we report average results over five independent runs.
As shown, the TKC adequacy criterion with the hyperparameters 
\textit{Gained}, $l(\overline{X^{ID}_{n}})<l(\overline{X^{OOD}_{n}})$,
and $HD \in (0.01-0.05]$ results in a coverage value that becomes lower as the number of TK neurons increases. 
Considering ResNet18, for instance,  coverage decreases from $82.69\%$ to $0.62\%$ for $TK_{10}$ ($\left\lvert TK_{10}\right\rvert=14$), and $TK_{50}$ ($\left\lvert TK_{50}\right\rvert=73$), respectively.
These results are not surprising, as the number of clusters of TK neurons extracted from equation~\eqref{eq::TCC} is between two and five. In fact, the decrease in \approach\ coverage reflects the rise in the combinations of TK neurons clusters (TCC) that increases exponentially as the percentage $p$ increases (e.g., $[86,104]$ for $\left\lvert TK_{10}\right\rvert=14$ and $[11994,1934664]$ for $\left\lvert TK_{50}\right\rvert=73$).

\begin{table}[t] 
\centering
\renewcommand{\arraystretch}{1.05}
\caption{Average TKC results for different percentages of TK neurons $p \in \{10\%, 20\%, 30\%, 40\%, 50\%\}$  
using Gained TK neurons and $HD \in (0.01-0.05]$}
\label{tab:par}
\begin{adjustbox}{width=\linewidth}
\Large
\begin{tabular}{l|llllll} 
\toprule
\multicolumn{2}{l}{\begin{tabular}[c]{@{}l@{}}$p\%$\\\end{tabular}} 
& \textbf{10} & \textbf{20} & \textbf{30}      & \textbf{40}      & \textbf{50}      \\ 

\hline
\multirow{2}{*}{\textbf{MNIST}}    & LeNet4                                                        & 75.00\% \scalebox{0.7}{$\left\lvert n\right\rvert=15$}                                            & 37.39\%                                          & 10.08\% & 6.43\%  & 0.43\%  \\ 
                          & LeNet1                                                        & 45.27\% \scalebox{0.7}{$\left\lvert n\right\rvert=17$}                                           & \begin{tabular}[c]{@{}l@{}}1.64\%\\\end{tabular} & 0.35\%  & 0.06\%  & 0.00\%  \\ 
\hline
\multirow{2}{*}{\textbf{SVHN}}     & LeNet5                                                        & \begin{tabular}[c]{@{}l@{}}70.12\% \scalebox{0.7}{$\left\lvert n\right\rvert=35$}\\\end{tabular} & 17.88\%                                          & 0.01\%  & 0.00\%  & 0.00\%  \\ 
                          & ResNet18                                                      & 38.69\% \scalebox{0.7}{$\left\lvert n\right\rvert=35$}                                           & 11.55\%                                          & 0.23\%  & 0.00\%  & 0.00\%  \\ 
\hline
\multirow{2}{*}{\textbf{CIFAR-10}} & LeNet5                                                        & 49.59\% \scalebox{0.7}{$\left\lvert n\right\rvert=15$}                                           & 41.67\%                                          & 19.82\% & 10.79\% & 0.13\%  \\ 
                          & ResNet18                                                      & 82.69\% \scalebox{0.7}{$\left\lvert n\right\rvert=14$}                                           & 38.76\%                                          & 25.57\% & 13.12\% & 0.62\%  \\
\bottomrule
\end{tabular}
\end{adjustbox}
\vspace{-4mm}
\end{table}

The results above provide evidence 
that TK neurons showing a steady increase in their features length $l$ during the ZeroShot stage, using the OOD dataset, can specialize their knowledge to the new domain, achieving a balance between overfitting and learning new knowledge in line with the generalisation principle~\cite{neyshabur2017exploring}.
Hence,
we conclude that \textbf{\approach\ is sensitive to its  hyperparameters, and is affected by the selected HD values, knowledge diversity, and the percentage of candidate TK neurons. Further, \approach\ benefits from choosing Gained TK neurons 
with $HD \in (0.01-0.05]$.}


\begin{table*}[ht]
\centering
\huge
\caption{\approach\ coverage (TKC) vs other metrics with adversarial inputs added to the original test set $S_{0}$}
\vspace{-1mm}
\label{ADV:effect}
\renewcommand{\arraystretch}{1.7}
\begin{adjustbox}{width=\textwidth}
\begin{tabular}{llllll|lllll|lllll|lllll|lllll}
 \hline
 & \multicolumn{5}{l|}{\textbf{LeNet1 (MNIST)}} & \multicolumn{5}{l|}{\textbf{LeNet4 (MNIST)}} & \multicolumn{5}{l|}{\textbf{LeNet5 (SVHN)}} & \multicolumn{5}{l|}{\textbf{ResNet18 (CIFAR-10)}} & \multicolumn{5}{l}{\textbf{LeNet5 (CIFAR-10)}} \\ \hline
    & \textbf{S$_{0}$} & \textbf{MIM} & \textbf{BIM} & \textbf{FGSM} & \textbf{PGD} 
    & \textbf{S$_{0}$} & \textbf{MIM} & \textbf{BIM} & \textbf{FGSM} & \textbf{PGD} 
    & \textbf{S$_{0}$} & \textbf{MIM} & \textbf{BIM} & \textbf{FGSM} & \textbf{PGD} 
    & \textbf{S$_{0}$} & \textbf{MIM} & \textbf{BIM} & \textbf{FGSM} & \textbf{PGD} 
    & \textbf{S$_{0}$} & \textbf{MIM} & \textbf{BIM} & \textbf{FGSM} & \textbf{PGD} 
    \\ \hline
\rowcolor[rgb]{0.953,0.953,0.953} \textbf{TKC$_{20}$} & 1.64\% & 3.49\% & 3.35\% & 3.43\% & 3.72\% & 37.39\% & 50.69\% & 51.11\% & 49.66\% & 50.80\% & 17.88\% & 19.10 & 18.43\% & 19.21\% & 19.08\% & 38.76\% & 45.07\% & 44.99\% & 45.14\% & 44.99\% & 41.67\% & 50.07\% & 49.80\% & 49.93\% & 49.80\% \\
\rowcolor[rgb]{0.953,0.953,0.953} \textbf{TKC$_{30}$} & 0.35\% & 0.73\% & 0.71\% & 0.72\% & 0.75\% & 10.08\% & 17.39\% & 17.30\% & 17.20\% & 16.88\% & 0.01\% & 0.04\% & 0.04\% & 0.07\% & 0.06\% & 25.57\% & 25.63\% & 25.57\% & 25.58\% & 25.57\% & 19.80\% & 19.95\% & 19.88\% & 19.85\% & 19.95\% \\
\rowcolor[rgb]{0.953,0.953,0.953} \textbf{TKC$_{40}$} & 0.06\% & 0.14\% & 0.11\% & 0.13\% & 0.13\% & 6.43\% & 12.17\% & 11.92\% & 12.13\% & 11.51\% & 0.00\% & 0.02\% & 0.01\% & 0.02\% & 0.02\% & 13.12\% & 13.15\% & 13.14\% & 13.14\% & 13.15\% & 10.79\% & 10.82\% & 10.84\% & 10.80\% & 10.82\% \\ \hline
\textbf{LSC} & 0.2\% & 0.3\% & 0.36\% & 4.0\% & 4.0\% & 0.20\% & 0.30\% & 0.40\% & 0.40\% & 0.40\% & 0.20\% & 0.30\% & 0.30\% & 0.30\% & 0.30\% & 0.3\% & 0.20\% & 0.20\% & 0.30\% & 0.30\% & 0.1\% & 0.26\% & 0.2\% & 0.3\% & 0.3\% \\
\textbf{KMNC} & 69.20\% & 71.92\% & 81.82\% & 83.60\% & 80.95\% & 62.71\% & 82.70\% & 79.60\% & 82.35\% & 84.70\% & 54.33\% & 71.60\% & 64.9\% & 62.7\% & 60.97\% & 43.60\% & 76.30\% & 64.40\% & 82.19\% & 69.28\% & 44.73\% & 59.7\% & 61.50\% & 66.11\% & 64.30\% \\
\textbf{NC} & 20.05\% & 20.05\% & 20.05\% & 20.05\% & 20.05\% & 23.80\% & 23.80\% & 23.80\% & 23.80\% & 23.80\% & 73.21\% & 73.57\% & 74.28\% & 77.85\% & 73.57\% & 71.78\% & 75.94\% & 67.21\% & 74.49\% & 73.76\% & 74.28\% & 75.03\% & 75.03\% & 75.03\% & 75.03\% \\
\textbf{NBC} & 7.97\% & 38.04\% & 40.94\% & 46.01\% & 41.66\% & 15.47\% & 46.42\% & 40.70\% & 47.61\% & 42.40\% & 22.67\% & 46.50\% & 45.30\% & 31.13\% & 23.57\% & 12.50\% & 28.10\% & 33.40\% & 41.30\% & 38.80\% & 9.10\% & 12.8\% & 20.3\% & 46.2\% & 40.7\% \\
\textbf{SNAC} & 12.31\% & 62.315 & 79.71\% & 73.18\% & 72.46\% & 19.04\% & 61.90\% & 53.40\% & 54.76\% & 54.76\% & 39.28\% & 13.80\% & 71.30\% & 83.20\% & 40.71\% & 21.42\% & 31.40\% & 37.20\% & 53.40\% & 45.10\% & 16.78\% & 21.1\% & 19.8\% & 41.3\% & 46.2\% \\
\textbf{TKNC} & 91.30\% & 91.30\% & 91.30\% & 91.30\% & 91.30\% & 100.00\% & 100.0\% & 100.00\% & 100.00\% & 100.00\% & 87.14\% & 88.21\% & 87.50\% & 87.50\% & 87.44\% & 87.85\% & 95.65\% & 94.92\% & 93.47\% & 95.65\% & 89.64\% & 89.64\% & 89.64\% & 98.8\% & 98.8\%\\ \hline
\end{tabular}
\end{adjustbox}
\end{table*}

\vspace{2mm}\noindent
\textbf{RQ3 (Effectiveness):}
The effectiveness of \approach\ as an adequacy criterion is evaluated by analysing its sensitivity to adversarial inputs and whether it can detect misbehaviours in datasets with inputs semantically different from the original test inputs.
Thus, we control input diversity by cumulatively adding inputs generated by different adversarial techniques. 
By exploiting the state-of-the-art adversarial approaches  FGSM~\cite{liu2019sensitivity}, BIM~\cite{kurakin2016adversarial}, MIM~\cite{dong2018boosting}, and PGD~\cite{deng2020universal}, multiple sets of adversarial inputs are devised with a size equal to 30\% of the original test set $S_{0}$.
Then, we execute the studied DNN models with these inputs and compare the observed changes of various coverage criteria, including  \approach. 
Similarly to state-of-the-art DNN testing criteria NC~\cite{pei2017deepxplore}, IDC~\cite{gerasimou2020importance}, KMNC, NBC, SNAC, TKNC~\cite{ma2018deepgauge}, and LSC~\cite{kim2019guiding}, we expect an increase in the computed \approach\ coverage.

Table \ref{ADV:effect} (rows $TKC_{20}$, $TKC_{30}$, $TKC_{40}$) reports the average \approach\ coverage results respectively for $p \in \left \{ 20\%, 30\%, 40\% \right\}$ across the three augmented test sets MNIST, SVHN and CIFAR-10 with the four different adversarial attacks.
The \approach\ coverage has a notable increase with LeNet1 and LeNet4 models trained with MNIST once adversarial examples are introduced. 
As shown in row $TKC_{20}$, \approach-based coverage increases for LeNet4 from $37.39\%$ for the original MNIST test inputs to  $51.11\%$ when BIM examples were introduced. Considering the CIFAR-10 dataset, the \approach\ values still increase when augmenting the test set with adversarial examples but with different margins for different DNNs. 
For ResNet18, as shown in row $TKC_{20}$ (Table~\ref{ADV:effect}), \approach increases by $\pm 7\%$ when introducing FGSM adversarial inputs compared to its original value with $S_{0}$. 

Compared to the original test set $S_{0}$, there is a considerable increase in the \approach\ result for the augmented test sets for all DNNs models. As expected, the \approach\ result for $p = 20\%$ ($TKC_{20}$) is higher than that for $p = 40\%$ ($TKC_{40}$) as the number of TCC grows exponentially with the number of TK neurons. 
In general, the increase is more significant in test sets involving adversarial inputs generated using MIM attacks.  
This study demonstrates the effectiveness of TK as a test adequacy criterion in detecting misbehaviours in test sets, revealing its sensitivity even when exposed to a small set of adversarial inputs.
Further, \approach\ is consistent with existing coverage criteria in measuring shifts in data distributions due to adversarial inputs~\cite{kim2019guiding,ma2018deepgauge}.

Based on these findings, \textbf{we conclude that \approach\ is sensitive to adversarial inputs
and is effective in helping software engineers detect misbehaviours in data and adversarial attacks in general.}

\vspace{2mm}\noindent
\textbf{RQ4 (Correlation):}
In this research question, we assess whether \approach\ is correlated to existing test adequacy criteria for DNNs.
We evaluate this correlation in response to semantically diverse input sets. 
In fact, we aim at a coverage metric that is consistent with existing  coverage criteria based on counting aggregation.
We control the input diversity by cumulatively adding inputs from the test set (i.e., $25\% S_{0}$, $50 \%S_{0}$, $75\% S_{0}$ and $100 \%S_{0}$). 
By gradually adding inputs, we generate sets that comprise diverse inputs with semantically different features. 
For instance, with over 26K inputs, the $25\% S_{0}$ of SVHN test set comprises less diverse features than $50\% S_{0}$.
We report results on how state-of-the-art coverage criteria perform across the three test sets SVHN, MNIST, and CIFAR-10 versus \approach\ in Table~\ref{corr}.

\begin{table*}[t]
\centering
\caption{\approach\ coverage (TKC) vs other metrics for different test set sizes (in \% of the original test set $S_0$)}
\vspace{-1mm}
\label{corr}
\begin{adjustbox}{width=\textwidth}
\renewcommand{\arraystretch}{1.2}
\begin{tabular}{l
>{\columncolor[HTML]{EFEFEF}}l 
>{\columncolor[HTML]{EFEFEF}}l 
>{\columncolor[HTML]{EFEFEF}}l 
>{\columncolor[HTML]{EFEFEF}}l llll
>{\columncolor[HTML]{EFEFEF}}l 
>{\columncolor[HTML]{EFEFEF}}l 
>{\columncolor[HTML]{EFEFEF}}l 
>{\columncolor[HTML]{EFEFEF}}l llll}
\hline
                 & \multicolumn{4}{c}{\cellcolor[HTML]{EFEFEF}\textbf{LeNet 4 (MNIST)}} & \multicolumn{4}{c}{\textbf{LeNet5 (SVHN)}}     & \multicolumn{4}{c}{\cellcolor[HTML]{EFEFEF}\textbf{LeNet5 (CIFAR-10)}} & \multicolumn{4}{c}{\textbf{ResNet18 (CIFAR-10)}} \\ \hline
\textbf{Test set size} 
    & \textbf{1/4 S}$_{0}$       
    & \textbf{1/2 S}$_{0}$         
    & \textbf{3/4 S}$_{0}$       
    & \textbf{S}$_{0}$           
    & \textbf{1/4 S}$_{0}$       
    & \textbf{1/2 S}$_{0}$         
    & \textbf{3/4 S}$_{0}$       
    & \textbf{S}$_{0}$           
    & \textbf{1/4 S}$_{0}$       
    & \textbf{1/2 S}$_{0}$         
    & \textbf{3/4 S}$_{0}$       
    & \textbf{S}$_{0}$           
    & \textbf{1/4 S}$_{0}$       
    & \textbf{1/2 S}$_{0}$         
    & \textbf{3/4 S}$_{0}$       
    & \textbf{S}$_{0}$           
    \\ \hline
 \textbf{TKC}$_{20}$           & 25.75\%       & 31.23\%       & 35.33\%      & 37.39\%      & 9.88\%  & 13.46\% & 15.94\% & 17.88\% & 40.62\%       & 40.62\%       & 41.67\%       & 41.67\%       & 28.00\%  & 34.64\%  & 36.87\% & 38.76\% \\ 
\textbf{TKC}$_{30}$           & 5.29\%        & 7.58\%        & 9.09\%       & 10.08\%      & 0.00\%  & 0.00\%  & 0.01\%  & 0.01\%  & 11.04\%       & 15.26\%       & 17.79\%       & 19.82\%       & 13.42\%  & 19.12\%  & 22.94\% & 25.57\% \\ 
\textbf{TKC}$_{40}$          & 2.88\%        & 4.34\%        & 5.55\%       & 6.43\%       & 0.00\%  & 0.00\%  & 0.00\%  & 0.00\%  & 5.22\%        & 7.71\%        & 9.45\%        & 10.79\%       & 6.68\%   & 9.67\%   & 11.65\% & 13.12\% \\ \hline
\textbf{IDC4}            & 12.50\%       & 12.50\%       & 12.50\%      & 12.50\%      & 12.50\% & 18.75\% & 18.75\% & 18.75\% & 25\%          & 25\%          & 25\%          & 25\%          & 27.67\%  & 27.67\%  & 27.67\% & 32.12\% \\ 
\textbf{IDC6}             & 3.12\%        & 3.12\%        & 3.12\%       & 3.12\%       & 3.12\%  & 4.69\%  & 4.69\%  & 4.69\%  & 10.94\%       & 10.94\%       & 10.94\%       & 10.94\%       & 20.94\%  & 20.94\%  & 24.04\% & 26.14\% \\ \hline
\textbf{LSC}              & 0.2\%         & 0.2\%         & 0.2\%        & 0.2\%        & 0.2\%   & 0.2\%   & 0.2\%   & 0.2\%   & 0.1\%         & 0.1\%         & 0.1\%         & 0.1\%         & 0.2\%    & 0.2\%    & 0.3\%   & 0.3\%   \\ 
\textbf{KMNC}             & 48.9\%        & 56.26\%       & 60.26\%      & 62.71\%      & 40.64\% & 47.49\% & 51.57\% & 54.33\% & 30.81\%       & 38.04\%       & 42.01\%       & 44.73\%       & 30.15\%  & 37.09\%  & 40.95\% & 43.60\% \\ 
\textbf{NC}               & 23.80\%       & 23.80\%       & 23.80\%      & 23.80\%      & 69.64\% & 71.42\% & 72.5\%  & 73.21\% & 68.57\%       & 71.42\%       & 72.85\%       & 74.28\%       & 68.21\%  & 70.0\%   & 71.42\% & 71.78\% \\ 
\textbf{NBC}              & 4.76\%        & 11.90\%       & 14.28\%      & 15.47\%      & 8.57\%  & 12.67\% & 18.92\% & 22.67\% & 1.78\%        & 4.82\%        & 7.14\%        & 9.10\%        & 2.32\%   & 6.60\%   & 10.35\% & 12.5\%  \\ 
\textbf{SNAC}             & 9.52\%        & 14.28\%       & 16.66\%      & 19.04\%      & 14.64\% & 22.85\% & 33.21\% & 39.28\% & 3.57\%        & 9.28\%        & 13.21\%       & 16.78\%       & 3.92\%   & 10.0\%   & 17.14\% & 21.42\% \\ 
\textbf{TKNC}             & 100.00\%      & 100.00\%      & 100.00\%     & 100.00\%     & 84.64\% & 87.14\% & 87.14\% & 87.14\% & 84.64\%       & 88.92\%       & 89.64\%       & 89.64\%       & 83.92\%  & 86.42\%  & 87.14\% & 87.85\% \\ \hline
\end{tabular}
\end{adjustbox}
\end{table*}

First, we found that TCC as a test adequacy criterion is consistent with the state-of-the-art coverage criteria based on neuron activation/property values such as NC~\cite{pei2017deepxplore}, KMNC~\cite{ma2018deepgauge} and NBC~\cite{ma2018deepgauge}. 
\approach's values increase with the increase of test set size, i.e., with MNIST, $p=20\%$ ($TKC_{20}$), the TKC score increases from $25.75\%$ to $ 31.23\%$ when cumulatively adding 25\% (column $1/2 S_{0}$) of the test set and reaches its maximum $37.39\%$ for the MNIST test set as soon as we reach 100\% of its size, i.e., $\left| S_{0} \right|=10K$.

An in-depth exploration of the obtained results reveals that \approach\ (rows $TKC_{20}$, $TKC_{30}$ and $TKC_{40}$ in Table~\ref{corr}) was able to detect these diversities through an increasing coverage value. 
In contrast, NC was stable with different sizes of $S_{0}$ for MNIST and exhibited a small increase with the other datasets.
On the other hand, coverage criteria based on input surprise, i.e., LSC, were stable when cumulatively increasing the size of the test set, e.g., 0.1\% for LeNet5 - CIFAR 10 and 0.2\% for LeNe4-MNIST. 
Since LSC~\cite{kim2019guiding} aims to measure the relative novelty/surprise of input with respect to the training set, it is legitimate to assume that the surprise level for these subsets of $S_{0}$ is insignificant and was not translated into an increase in the LSC score as the test set size increases. This is an interesting finding that requires further investigation.
While the original tests $100\% S_{0}$ achieves 62.71\% KMNC (k = 10) on LeNet4, it only obtains 48.9\% on $25\%$ of  $S_{0}$. In line with \approach\ and KMNC results, the $100\%$ of $S_{0}$ MNIST test set achieves a higher value for NBC with $15.47\%$ while $25\%$ of $S_{0}$ achieves only 4.76\%. 
We can notice the same behaviour with all the DNN and dataset combinations we have tested. 
The observed trend supports our claim that \approach\ assesses an input based on the knowledge (i.e., features) that can be abstracted from it by the TK neurons. 
Hence, the TK neurons learn unique features from within a dataset based on how well the input data can cover various TCC combinations ~\eqref{eq::TCC}, which ensures the semantic diversity of the tested inputs. 


Overall, \approach\ coverage has a good performance in detecting the diversity of the test set. 
TKC increases with the increase of test set size, proving that the \approach\ adequacy criterion is sound~\cite{pezze2008software} and sensitive to semantically different inputs. 
Further, \approach\ coverage shows similar behaviour to state-of-the-art coverage criteria for DNN systems, and is consistent with coverage criteria based on aggregating neuron property values (e.g., NBC, IDC).

Based on these findings, \textbf{we confirm the positive correlation between our approach and the studied neuron coverage criteria. We also conclude that the TKC test adequacy criterion can support software engineers to evaluate the diversity of a test set that comprises semantically different test inputs.}

\vspace{2mm}\noindent
\textbf{RQ5 (OOD Selection).}
\approach\ is mainly based on suitable ID and OOD dataset selection. 
Thus, we examine the robustness of our approach by evaluating the effect of ID/OOD dataset combinations on the \textit{TK neurons selection} stage of \approach\ as well as the final coverage estimation.
For each model, we replicate the experiments with the same settings and hyperparameters, changing only the OOD dataset. 
We monitor the OOD selection based on the rules discussed earlier in Section~\ref{sec:approach}, so that the ID and OOD data have common features, i.e., color, resolution, etc.

Table~\ref{tab:OODc} illustrates the results where we measure the consistency of the coverage against the changes in the OOD dataset. 
The results reveal that on average the number of TK neurons correlates strongly with the type of OOD data. 
Upon closer examination, we observe that the variations occur not only because of increasing the number of TK, but also due to changing most of them when the OOD dataset changes.
Thus, the alteration in TK neurons has impacted the coverage estimation. For instance, the coverage for LeNet1-MNIST with adequacy criterion 30\% of TK neurons ($TKC_{30}$) increased from $30.23\%$  to $71.43\%$ when using EMNIST as the OOD dataset instead of Fashion MNIST, although the size of TK set has been reduced from 29 to 12. 
We observe a similar pattern with LeNet4-MNIST when we deploy EMNIST as the OOD dataset.
With LeNet5-SVHN the change in OOD data displays unexpected behaviours. We observe that both the number of TK neurons and the coverage decreased when the EMNIST OOD data was replaced by the GTSRB dataset.
Independently of the number of TK neurons, the coverage increases based on the covered TCC combinations. In fact, when equipped with a suitable OOD dataset, we can effectively identify TK neurons that capture the most essential semantic features present in the dataset, i.e, coverage combinations TCC. This, in turn, leads to a more equitable and accurate coverage estimation when testing the model on the test set.

Therefore, to ensure a comprehensive testing process, it is essential to limit the selection of ID/OOD combinations to datasets where all images belong to the same ``macro'' category. This restriction enables inputs to share crucial semantic features, facilitating the identification of the appropriate TK neurons.
Based on these findings, \textbf{we conclude that the selection of OOD dataset does affect the whole \approach\ workflow and the coverage estimation in particular.}

\begin{table}[t]
\renewcommand{\arraystretch}{1.05}
\small
\centering
\begin{adjustbox}{width=\columnwidth}
\begin{threeparttable}
\caption{Comparison of different OOD datasets impact on the number of TK neurons and TKC score}
\label{tab:OODc}
\vspace{-2mm}
\begin{tabular}{lccccll}
\hline
                & \multicolumn{2}{c}{\textbf{LeNet1 (MNIST)}}                              & \multicolumn{2}{c}{\textbf{LeNet4 (MNIST)} }                              & \multicolumn{2}{c}{\textbf{LeNet5 (SVHN)}}                               \\ \hline
            \textbf{OOD}             
                & \textbf{F.MNIST}                  
                & \textbf{EMNIST}                         
                & \textbf{F.MNIST}                  
                & \textbf{EMNIST}                         
                & \textbf{EMNIST}     
                & \textbf{GTSRB}      
                \\ \hline
        \textbf{\#TK} 
                & 29                             
                & 12                             
                & 24                             
                & 12                             
                & 87                             
                & 75                             \\
\textbf{TKC}$_{30}$           & {\color[HTML]{333333} 30.23\%} & {\color[HTML]{333333} 71.43\%} & {\color[HTML]{333333} 28.89\%} & {\color[HTML]{333333} 66.67\%} & {\color[HTML]{333333} 39.71\%} & {\color[HTML]{333333} 14.01\%} \\
\textbf{TKC}$_{100}$         & {\color[HTML]{333333} 38.32\%} & {\color[HTML]{333333} 49.44\%} & {\color[HTML]{333333} 34.30\%} & {\color[HTML]{333333} 50.13\%} & {\color[HTML]{333333} 8.71\%}  & {\color[HTML]{333333} 1.37\%}  \\ \hline
\end{tabular}
\begin{tablenotes}
\small
\item \textcolor{darkgray}{*OOD dataset size: 800 samples for all model/data combinations. F.MNIST: Fashion MNIST dataset; \#TK: total number of TK neurons}.
\end{tablenotes}
\end{threeparttable}
\end{adjustbox}
\vspace{-5mm}
\end{table}

\vspace{-1mm}
\subsection{Threats to Validity \label{Threats}}
\vspace{-1mm}
\noindent
\textbf{Construct Validity.}
The primary threat to the construct validity of \approach\ is the correctness of the studied DNN models and the  datasets used.
We mitigate these construct validity threats using widely-studied datasets, and also employing publicly-available architectures and pre-trained DNN models that achieved competitive performance; see Table~\ref{table:Dl_data}. 
Further, we mitigate threats to identifying core units responsible for the DNN prediction and knowledge abstraction by adapting the state-of-the-art approach for the analysis of knowledge generalisation under domain shift~\cite{kawaguchi2017generalization,rethmeier2020tx}.

\vspace{0.5mm}\noindent
\textbf{Internal Validity.}
The main internal validity threat that could introduce bias to \approach\ is the computation of the TKC score. 
To mitigate this threat, we design a study based on a specific set of research questions. 
We first illustrate the effectiveness of the \approach\ selection methodology of TK neurons and their effectiveness as a test adequacy criterion in RQ1 and RQ4. 
For RQ4, we illustrate \approach's effectiveness by augmenting the original test sets with numerically diverse inputs. 
RQ3 was also designed to illustrate the effectiveness of \approach\ to detect adversarial examples and confirmed its positive correlation with the state-of-the-art test adequacy criteria NC~\cite{pei2017deepxplore}, KMNC, NBC, SNAC, TKNC~\cite{ma2018deepgauge}, LSC~\cite{kim2019guiding} and IDC~\cite{gerasimou2020importance}. 
The performance of our approach is also demonstrated through RQ2 for different values of TK neurons $p \in \left \{ 10\%, 20\%, 30\%, 40\%, 50\% \right \}$, Hellinger Distance $HD \in (0.01-0.05]$ and neuron types (`Gained' or `Stable'), signifying knowledge change and knowledge diversity, respectively~\cite{rethmeier2020tx}. 

\vspace{0.5mm}\noindent
\textbf{External Validity.}
Threats to external validity have been mostly mitigated by developing \approach\ on top of the open-source frameworks Keras and Tensorflow. We also validated \approach\ using popular DNNs trained on publicly-available datasets (e.g., MNIST, SVHN, CIFAR-10). 
However, more experiments are needed to validate \approach\ for TK neurons' extraction using Supervised transfer learning rather than ZeroShot transfer learning. This will enable us to assess if TK neurons can generalise knowledge in more challenging domain shift scenarios.

\vspace{-1mm}
\section{Related Work\label{sec:relatedwork}}
\vspace{-1mm}
Testing and assurance of DNN-based systems is a growing research field~\cite{BurtonGH2017,singh2019abstract}. 
Recent research efforts have resulted in a broad range of approaches for DNN testing; see~\cite{zhang2020testing,huang2020survey} for recent surveys. 
In fact, DNN behaviour heavily relies on the training data and learned parameters such as weights and biases, which makes it impractical to apply conventional software testing techniques to ensure the quality and safety of DNNs.
%
Thus, recent advances introduced techniques for DNN adequacy criteria~\cite{pei2017deepxplore,ma2018deepgauge,gerasimou2020importance,kim2019guiding}, input generation and selection~\cite{tian2018deeptest,ma2021test,sun2020automatic}, fault localisation~\cite{eniser2019deepfault} and repair~\cite{sohn2023arachne,kim2023repairing}.

Test adequacy, in particular, provides a quantitative assessment of the effectiveness of a testing set, where achieving high coverage indicates comprehensive testing of the model~\cite{zhang2020machine}. 
DeepXplore~\cite{pei2017deepxplore} pioneered research by proposing neuron coverage (NC) as a white-box testing criterion to identify the parts of DNN logic exercised by a test set.  Simply put, NC measures the ratio of neurons whose activation values are above a predefined threshold. Eventually, it helps software testers to select properties of a program to focus on during testing.
More fine-grained approaches were introduced which provide improvements over NC, including DeepGauge~\cite{ma2018deepgauge}, DeepGini~\cite{feng2020deepgini}, DeepTest~\cite{tian2018deeptest} and DeepImportance~\cite{gerasimou2020importance}. 
For instance, DeepGauge~\cite{ma2018deepgauge} has introduced multi-granularity testing criteria based on a more detailed analysis of neuron activation values. 
It defines an activation interval that is divided into $k$ equal sections, and the coverage of the test set is estimated based on its ability to cover these sections. 
Similarly, recent studies explored the computational behaviour of DNNs by looking at each neuron's contribution to the model's decision-making. For instance, DeepImportance~\cite{gerasimou2020importance} leverages an explanation method~\cite{bach2015pixel} to evaluate how prominent each neuron and its connections are in the model's final prediction through relevance propagation techniques. 

Concerns about the DNN stability to adversarial attacks were also investigated~\cite{dong2018boosting,gopinath2017deepsafe} with  approaches proposing solutions to optimize the DNN performance in response to malicious data perturbations.
DeepHunter~\cite{xie2019deephunter} proposes a data augmentation strategy as a possible solution.
Another line of research investigated more general correctness properties related to data distributions. 
A general framework for automated verification of safety DNN properties is proposed in~\cite{huang2017safety}. The authors assess the safety at an individual decision level in terms of invariance of the classification within a specific region of a data point, enabling the propagation of this analysis layer by layer, thus allowing greater scalability.
Similarly, SafeML~\cite{aslansefat2020safeml}, like LSC~\cite{kim2019guiding}, addresses the distribution shift problem by exploiting statistical distance measures. The technique compares training datasets versus the inference data input to assess the DNN robustness independently of the training approach itself. 

Most developed testing techniques extend existing software testing techniques via simple adaptations to DNN. However, none of them investigated how knowledge transfer and knowledge generalisation~\cite{urolagin2011generalization} can be used to inform test adequacy analysis. 
In contrast, \approach\ drives the identification of transfer knowledge neurons based on knowledge change and diversity. 


\vspace{-2mm}
\section{Conclusions and Future Work\label{sec:conclusions}}
\vspace{-1mm}
\approach\ targets DNN testing by quantifying the knowledge diversity of a test set. 
Our approach leverages the generalisation property of DNNs to analyze the knowledge transfer capability of a DNN under domain shift. 
Thus, it identifies a finite set of neurons responsible for knowledge generalisation and uses this information to define the Transfer Knowledge Coverage (TKC) adequacy criterion for assessing how well the test set exercises different DNN behaviours. 
Our empirical evaluation shows the effectiveness of \approach\ in improving DNN's accuracy when used to guide the selection of inputs to augment the training dataset. 
Our future work involves 
(i) extending \approach\ and the TKC test adequacy criterion to support object detection models (e.g., YOLO~\cite{redmon2016you}); 
(ii) automating  data augmentation using \approach\ to reduce the cost of data creation and labelling; and 
(iii) adapting \approach\ to support model pruning to improve the DNN robustness and accuracy.

\bibliographystyle{elsarticle-harv} 
\bibliography{main}

\end{document}